\documentclass{article}

\usepackage[final]{neurips_2019}

\usepackage[utf8]{inputenc} 
\usepackage[T1]{fontenc}    
\usepackage{hyperref}       
\usepackage{url}            
\usepackage{booktabs}       
\usepackage{amsfonts}       
\usepackage{nicefrac}       
\usepackage{microtype}      
\usepackage{array}
\usepackage{makecell}
\usepackage{xcolor}
\usepackage{graphicx}

\title{Zero-shot Text Classification With Generative Language Models}

\author{
  Raul Puri \\
  NVIDIA\\
  \texttt{raulp@nvidia.com} \\
  \And
  Bryan Catanzaro \\
  NVIDIA\\
  \texttt{bcatanzaro@nvidia.com} \\
}

\begin{document}

\maketitle

\begin{abstract}
   This work investigates the use of natural language to enable zero-shot model adaptation to new tasks. We use text and metadata from social commenting platforms as a source for a simple pretraining task. We then provide the language model with natural language descriptions of classification tasks as input and train it to generate the correct answer in natural language via a language modeling objective. This allows the model to generalize to new classification tasks without the need for multiple multitask classification heads. We show the zero-shot performance of these generative language models, trained with weak supervision, on six benchmark text classification datasets from the torchtext library. Despite no access to training data, we achieve up to a 45\% absolute improvement in classification accuracy over random or majority class baselines. These results show that natural language can serve as simple and powerful descriptors for task adaptation. We believe this points the way to new metalearning strategies for text problems.
  
\end{abstract}

\section{Method}
Our method reformulates text classification problems as multiple choice question answering. To enable our model to generalize to new classification tasks, we provide the model with a multiple choice question description containing each class in natural language, and train it to generate the correct answer, also in natural language, from the provided description. To better prepare our model to handle a wide variety of class descriptors, we utilize a pretrained GPT-2 \citep{Radford2019GPT2} transformer model and finetune it on the task of multiple choice title prediction for the OpenWebText dataset \citep{openwebtext}. This pretraining task trains the model to use common sense reasoning to select the most probable title or description of the text data from a provided list of rich natural language descriptions or classes, similar to the problem formulation of text classification. The wide variety of titles available in the pretraining dataset help simulate numerous automatically generated \textit{N}-way text classification tasks to enable meta-learning.
In initial studies we found that the diverse language found in title prediction was necessary to adapt to new tasks, and other pretraining tasks such as WebText subreddit prediction did not transfer at all.

For a given document, we randomly sample a number of titles $t \in [2, 15]$ with one title being the correct title. Half of the time we replace a single title with “none of the above”, and occasionally ($p=1/t$) we choose to replace the correct title with “none of the above”. We prepend all selected titles to the document in the form of a multiple choice question, and train the model to generate the answer, similar to generative Question Answering \citep{mqan}. Example input representations for title prediction can be found in Table \ref{tab:example_data}.

The model is optimized by computing a next token prediction language modeling loss, $\sum_t \mathcal{L}(w_t, P(\hat{w}_t|w_{[1, t-1]}))$,  that optimizes over the entire concatenated input $w=[question,$ $ reference\_text, output\_answer]$ and the questions are generated according to a grammar. The input representation utilizes type tokens to segment the question, reference text, and answer. To encode positional information, the input uses learned positional embeddings that reset to position 0 at the start of the answer. This is described in more detail in the appendix section \ref{mcq}. 
\begin{figure*}[t!]
	\begin{center}
		\includegraphics[scale=.5]{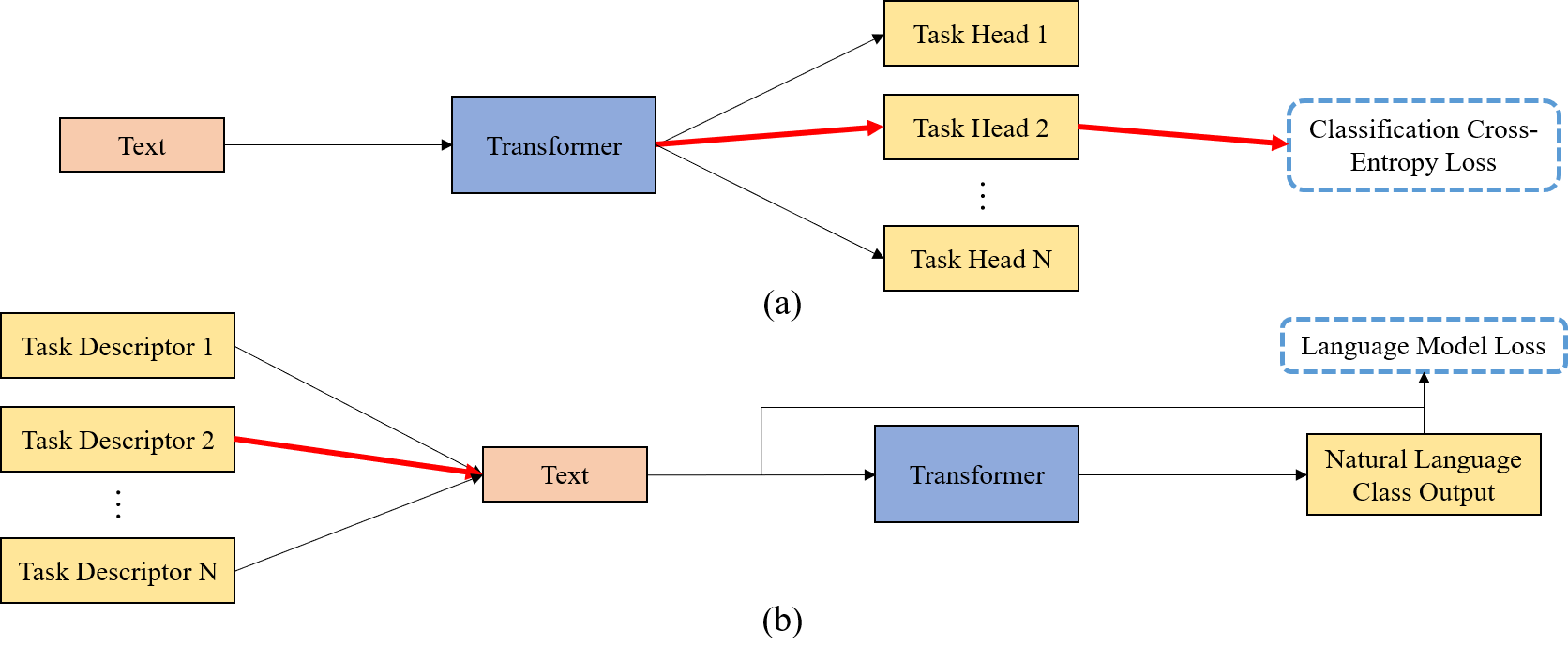}
		\caption{Comparison between existing multitask classifiers and our method. (a) Multitask classifiers have the model featurize text and send it to one of N task heads. (b) In our method, one of N task descriptors is prepended to the text and the model generates the answer in natural language.}
		\label{fig:confusions}
	\end{center}
\end{figure*}
\begin{table*}[bt!]
\footnotesize
\begin{center}
\label{tab:example_data}
\begin{tabular}{|p{2cm}|p{4.1cm}|p{4.1cm}|p{1.8cm}|}
\hline 
\thead{Dataset} & \thead{Question} & \thead{Text} & \thead{Answer} \\ \hline
Title Prediction Pretraining & Which of these choices best describes the following document? : `` A pool For All Bodies  '' , `` Lawmakers say they'd take pay cut, but they can't  '' , `` Raiders' Gareon Conley faces civil suit '' ,  `` Prolific cybercriminal suspected of spreading ransomware arrested by Polish Police [Europol]  '' &
Story highlights Members of Congress also preparing for potential sharp cuts in federal spending\textbackslash n\textbackslash nBut lawmakers will not see any change to their annual salary of \$174,000... &
Lawmakers say they'd take pay cut, but they can't \\\hline
AGNews Zero-shot Classification & How is the text best described? :  `` Science \& Technology '' , `` Business '' , `` Sports '' , or `` World News '' & An Entertaining Holiday Pick \textbackslash n
Hastings, a multimedia retailer, trims losses and raises full-year guidance. & Business \\\hline

\end{tabular}
\caption{Example inputs for pretraining and downstream tasks. The descriptor questions are concatenated to the text samples and the language model generates the remaining output answer text. Class descriptors for the 5 other downstream classification tasks can be found in appendix \ref{class_description}}
\end{center}
\end{table*}

For our analysis of zero shot classification we examine the performance of our model at various sizes on several of the torchtext classification datasets. When transferring the model we provide all the given dataset’s classes (typically ranging from 2-15 classes) to the model in the multiple choice question format and prompt it to generate out the correct class. Furthermore, we ensure that downstream tasks do not contain "none of the above" options. We use greedy autoregressive decoding to generate our output text. Example inputs for each of our downstream tasks are shown in Table \ref{tab:example_data}.

\subsection{Dataset}
We build upon prior work collecting large language modeling datasets from the internet. Namely, we extend the OpenWebText corpus \citep{openwebtext} by annotating the documents with subreddits and titles in natural language. The OpenWebText dataset is collected by scraping outbound weblinks from reddit that have more than 3 karma score. We annotate each outbound weblink with the title of the Reddit post, and the subreddit that the link was posted in. Weblinks can appear in multiple posts across different subreddits, so for a given link we aggregate a list of all it’s related subreddits and titles. Detailed dataset statistics can be found in appendix section \ref{data_statistics}.
To create training data we sample a random document, multiple titles including one of the documents corresponding titles, and arrange the input as described in the previous section. We evaluate the trained model on the DBPedia, AGNews, Yahoo Answers, SST-2, Amazon-2, and Yelp-2 text classification datasets \citep{sst2,dbpedia}. The classes and class descriptors used for each of these tasks can be found in appendix section \ref{class_description}. To experiment with different class descriptions and model architectures, we create a small validation set of 2000 random training set examples for each of the downstream tasks. We evaluate our design choices on these validation sets before reporting final accuracies on the entire test set.

\section{Related Work}

Zero and few shot learning have been the subject of many studies. Some works have looked at meta-learning for machine translation in low resource languages \citep{ml_nmt}, iteratively guiding policies with language \citep{nlp_guided_ml} for instruction following \citep{branavan2009reinforcement,chen2011learning}, and generating WikiSQL-style structured queries from natural language queries \citep{structured_query}. \cite{Radford2018GPT,Radford2019GPT2} show that large scale language models can be used in a multitask zero shot capacity by allowing the model to generate output text in an autoregressive manner given a prompt with the task description. They demonstrate that larger transformer language models perform better than smaller models in zero shot settings. However, their models are never explicitly trained for zero shot text classification. To perform classification, the authors propose appending a prompt token to the text and restricting the output vocabulary to the tokens of possible answers. This effectively turns the output vocabulary into a pretrained task-specific classification head. Unlike our approach their work requires manual intervention and does not take advantage of task descriptors to modulate output behavior. The Multitask Question Answering Network \citep{mqan} study also investigates zero shot performance of multitask generative language models prompted with descriptor questions. However, they only analyze zero shot classification performance between tasks of identical domains (SST-2 and Amazon-2) that are trained with supervised learning and identical prompts. Using identical prompts and supervised learning prevents a true analysis of the model's ability to adapt to unseen task descriptors.

Recent work in meta-learning has centered around gradient based meta learning strategies such as Model Agnostic Meta-Learning or MAML \citep{maml}. However, parallel work such as Memory Augmented Neural Networks \citep{MANN} and Simple Neural Attentive Learners \citep{SNAIL} demonstrate the effectiveness of architecture based meta-learning. This is similar to our work except that our models receive weak supervision in the form of class labels and a question in natural language instead of similar class examples. We show throughout this work that melding techniques from NLP and architecture based meta-learning allows our model to adapt to new language classification tasks.

Lastly, similar to our work, concurrent research investigates models capable of handling tasks with different class counts and output mappings. \cite{multifs} combine prototypical networks and MAML to adapt to NLP tasks with different numbers of labels. \cite{T5} propose a unified multitask language model that uses weakly-supervised task labels to generate task outputs with natural language. By doing so, the resulting model is capable of performing a diverse set of tasks including classification, natural language inference, question answering, and abstractive summarization.  Furthermore, the authors demonstrate the viability of this approach by scaling the model to 11 billion parameters and achieving state of the art accuracy. However, neither of these works examine the ability of a unified model to adapt to new task descriptors in a zero-shot fashion.

\section{Results}

\begin{table}[t!]
\footnotesize
\begin{center}

\begin{tabular}{|c|c|c|c|c|c|c|c|}
\hline
Model & SST-2 & AGNews & DBPedia & Yahoo & Amazon-2 & Yelp-2 & Average \\ \hline
Random Guess\textsuperscript{\textasciitilde} & 50.6 & 27.4 & 7.27 & 10.2 & 52.9  & 50.4
& 33.1 \\\hline
Majority Class\textsuperscript{\textasciitilde} & 49.9 & 25.3 & 7.6 & 9.9 & 49.3 & 49.2 & 31.9 \\\hline\hline
117M All Data & 51.8 / 0 & 40.2 / .00 & 39.6 / .25 & 26.1 / .97 & 50.3 / .001 & 50.1 / 0 & 43.0 / .202 \\ \hline 
355M 1/4 Data & 61.7 / 0 & \textbf{68.3 / .51} & \textbf{52.5 / .03} & \textbf{52.2 / .64} & 64.5 / .001 & 58.5 / 0 & 59.6 / .197 \\ \hline
355M All Data & \textbf{62.5 / 0}  & 65.5 / .01 & 44.8 / .62 & 49.5 / .30 & \textbf{80.2 / 0} & \textbf{74.7 / 0} & \textbf{62.9 / .176} \\ \hline\hline
355M Finetuned\textsuperscript{\textasciitilde} & 93.23	& 94.87 &	99.0	& 72.79	& 97.115 & 94.479 & 91.91 \\\hline 
SOTA & 96.8* & 95.51* & 99.38* & 76.26** & 97.6* & 98.45* & 94 \\\hline

\end{tabular}
\label{tab:zero_shot_results}
\caption{Zero shot transfer results. Seperated by a slash, each column contains test accuracies and (when applicable) the percentage of out of vocabulary test answers. Provided baseline models include random guessing\textsuperscript{\textasciitilde}, majority class\textsuperscript{\textasciitilde}, and finetuning\textsuperscript{\textasciitilde} baselines. State of the art results held by *XLNet \citep{XLNet} and **DRNN \citep{DRNN}.}

\end{center}
\end{table}

To test the ability of our pretrained models to adapt to new tasks and tasks descriptions, we transfer the models to 6 classification tasks. We provide three baselines the first two of which are designed to expose dataset bias: random guessing, majority class (mode of the training dataset), and directly finetuning a 355 million parameter classification model on the downstream tasks. In our experiments we investigate the effect of two components of the pretraining process on downstream task performance: model scale and data scale. Table 2 shows that increasing model size leads to improved performance on downstream tasks. In some scenarios smaller models are barely able to perform better than random. For DBPedia the 355M GPT-2 model leads to a 45.2\% absolute accuracy improvement over random. In tasks with several classes such as DBPedia, AGnews, and Yahoo Answers the model performs noticeably better than random; however, they struggle to break past 50\% and no task comes close to achieving either finetuned or SOTA accuracies. Contextualizing these results with the results of the binary classification tasks like SST-2, Amazon-2, and Yelp-2 we hypothesize that the model can narrow down unlikely classes, but struggles to choose between the two most plausible options due to its lack of formal supervision. 

These results also show that restricting the size of the dataset and available document-title pairs leads to a reduction in overall task performance averaged across all tasks. This highlights the need for pretraining across a diverse set of tasks and language. Table 2 demonstrates that the robustness of our generative model is also similarly dictated by model and pretraining dataset size. Although rare across all pretrained models, the out of distribution answers (generated answers that are not valid classes) diminish with larger pretrained models and data. The most common out of vocab answer is an empty string where the model decides to immediately predict the end of text token. Other out of vocab answers are typically rearrangements of valid answer tokens. These are rare with greedy decoding, but become more frequent when using other sampling methods such as top-k \citep{topk} or top-p nucleus sampling \citep{topp}. In the case of Yahoo Answers the model can combine two categories such as "Education \& Reference" with "Science \& Mathematics" to output "Education \& Mathematics". We perform further studies examining the relationship between question descriptions, tokenization, accuracy, and out of vocabulary answers in appendix section \ref{qsec}. These studies showcase the model's ability to adapt to different descriptions, but expose issues with controllability. Nevertheless, with this model the practitioner's burden is shifted away from designing effective zero-shot multitask architectures, to data problem design.

\section{Conclusion and Future Work}
In this work, we present a novel pretraining method for zero shot language classification through a generative language model classifier. By generating classifications through natural language, the model eliminates the need for multiple task-specific classification heads, making the model far more general and flexible. Increasing model and data scale further demonstrates that the capabilities of recent transformer language models are sufficient to extract meaningful feature representations that allow us to better generalize and adapt to new tasks. These results highlight the potential of natural language as learning and adaptation signals in future applications.

Currently this work is employed for zero-shot classification. Future extensions should investigate the ability of gradient based metalearning to adapt to task descriptors, either through K-shot support-based learning or by taking gradient steps on the task descriptors themselves as in \cite{updaterules}. Additionally, future work could extend the text classification task to other language problems such as question answer or instruction following. Applying this technique in other settings will require addressing its current limitations with respect to controllability, available data and task diversity.

\bibliographystyle{humannat}
\bibliography{neurips_2019}

\begin{thebibliography}{}

\bibitem[\protect\astroncite{Bansal et~al.}{2019}]{multifs}
Bansal, T., R.~Jha, and A.~McCallum\leavevmode\nopagebreak\newline 2019.
\newblock Learning to few-shot learn across diverse natural language
  classification tasks.
\newblock {\em arXiv preprint arXiv:1911.03863}.

\bibitem[\protect\astroncite{Branavan et~al.}{2009}]{branavan2009reinforcement}
Branavan, S.~R., H.~Chen, L.~S. Zettlemoyer, and
  R.~Barzilay\leavevmode\nopagebreak\newline 2009.
\newblock Reinforcement learning for mapping instructions to actions.
\newblock In {\em Proceedings of the Joint Conference of the 47th Annual
  Meeting of the ACL and the 4th International Joint Conference on Natural
  Language Processing of the AFNLP: Volume 1-Volume 1}, Pp.~ 82--90.
  Association for Computational Linguistics.

\bibitem[\protect\astroncite{Chen and Mooney}{2011}]{chen2011learning}
Chen, D.~L. and R.~J. Mooney\leavevmode\nopagebreak\newline 2011.
\newblock Learning to interpret natural language navigation instructions from
  observations.
\newblock In {\em Twenty-Fifth AAAI Conference on Artificial Intelligence}.

\bibitem[\protect\astroncite{Co{-}Reyes et~al.}{2018}]{nlp_guided_ml}
Co{-}Reyes, J.~D., A.~Gupta, S.~Sanjeev, N.~Altieri, J.~DeNero, P.~Abbeel, and
  S.~Levine\leavevmode\nopagebreak\newline 2018.
\newblock Guiding policies with language via meta-learning.
\newblock {\em CoRR}, abs/1811.07882.

\bibitem[\protect\astroncite{Fan et~al.}{2018}]{topk}
Fan, A., M.~Lewis, and Y.~N. Dauphin\leavevmode\nopagebreak\newline 2018.
\newblock Hierarchical neural story generation.
\newblock {\em CoRR}, abs/1805.04833.

\bibitem[\protect\astroncite{Finn et~al.}{2017}]{maml}
Finn, C., P.~Abbeel, and S.~Levine\leavevmode\nopagebreak\newline 2017.
\newblock Model-agnostic meta-learning for fast adaptation of deep networks.
\newblock {\em CoRR}, abs/1703.03400.

\bibitem[\protect\astroncite{Gu et~al.}{2018}]{ml_nmt}
Gu, J., Y.~Wang, Y.~Chen, K.~Cho, and V.~O.~K.
  Li\leavevmode\nopagebreak\newline 2018.
\newblock Meta-learning for low-resource neural machine translation.
\newblock {\em CoRR}, abs/1808.08437.

\bibitem[\protect\astroncite{Holtzman et~al.}{2019}]{topp}
Holtzman, A., J.~Buys, M.~Forbes, and Y.~Choi\leavevmode\nopagebreak\newline
  2019.
\newblock The curious case of neural text degeneration.
\newblock {\em CoRR}, abs/1904.09751.

\bibitem[\protect\astroncite{Huang et~al.}{2018}]{structured_query}
Huang, P., C.~Wang, R.~Singh, W.~Yih, and X.~He\leavevmode\nopagebreak\newline
  2018.
\newblock Natural language to structured query generation via meta-learning.
\newblock {\em CoRR}, abs/1803.02400.

\bibitem[\protect\astroncite{Kingma and Ba}{2014}]{Adam}
Kingma, D.~P. and J.~Ba\leavevmode\nopagebreak\newline 2014.
\newblock Adam: A method for stochastic optimization.
\newblock {\em arXiv preprint arXiv:1412.6980}.

\bibitem[\protect\astroncite{Lehmann et~al.}{2015}]{dbpedia}
Lehmann, J., R.~Isele, M.~Jakob, A.~Jentzsch, D.~Kontokostas, P.~N. Mendes,
  S.~Hellmann, M.~Morsey, P.~Van~Kleef, S.~Auer,
  et~al.\leavevmode\nopagebreak\newline 2015.
\newblock Dbpedia--a large-scale, multilingual knowledge base extracted from
  wikipedia.
\newblock {\em Semantic Web}, 6(2):167--195.

\bibitem[\protect\astroncite{Loshchilov and
  Hutter}{2019}]{loshchilov2018decoupled}
Loshchilov, I. and F.~Hutter\leavevmode\nopagebreak\newline 2019.
\newblock Decoupled weight decay regularization.
\newblock In {\em International Conference on Learning Representations}.

\bibitem[\protect\astroncite{McCann et~al.}{2018}]{mqan}
McCann, B., N.~S. Keskar, C.~Xiong, and
  R.~Socher\leavevmode\nopagebreak\newline 2018.
\newblock The natural language decathlon: Multitask learning as question
  answering.
\newblock {\em CoRR}, abs/1806.08730.

\bibitem[\protect\astroncite{Metz et~al.}{2018}]{updaterules}
Metz, L., N.~Maheswaranathan, B.~Cheung, and
  J.~Sohl-Dickstein\leavevmode\nopagebreak\newline 2018.
\newblock Meta-learning update rules for unsupervised representation learning.
\newblock {\em arXiv preprint arXiv:1804.00222}.

\bibitem[\protect\astroncite{Micikevicius et~al.}{2017}]{MPTraining}
Micikevicius, P., S.~Narang, J.~Alben, G.~F. Diamos, E.~Elsen, D.~Garcia,
  B.~Ginsburg, M.~Houston, O.~Kuchaiev, G.~Venkatesh, and
  H.~Wu\leavevmode\nopagebreak\newline 2017.
\newblock Mixed precision training.
\newblock {\em CoRR}, abs/1710.03740.

\bibitem[\protect\astroncite{Mishra et~al.}{2017}]{SNAIL}
Mishra, N., M.~Rohaninejad, X.~Chen, and
  P.~Abbeel\leavevmode\nopagebreak\newline 2017.
\newblock Meta-learning with temporal convolutions.
\newblock {\em CoRR}, abs/1707.03141.

\bibitem[\protect\astroncite{Peterson et~al.}{2019}]{openwebtext}
Peterson, J., S.~Meylan, and D.~Bourgin\leavevmode\nopagebreak\newline 2019.
\newblock Open clone of openai's unreleased webtext dataset scraper.

\bibitem[\protect\astroncite{Radford et~al.}{2018}]{Radford2018GPT}
Radford, A., K.~Narasimhan, T.~Salimans, and
  I.~Sutskever\leavevmode\nopagebreak\newline 2018.
\newblock Improving language understanding by generative pre-training.

\bibitem[\protect\astroncite{Radford et~al.}{2019}]{Radford2019GPT2}
Radford, A., J.~Wu, R.~Child, D.~Luan, D.~Amodei, and
  I.~Sutskever\leavevmode\nopagebreak\newline 2019.
\newblock Better language models and their implications.

\bibitem[\protect\astroncite{Raffel et~al.}{2019}]{T5}
Raffel, C., N.~Shazeer, A.~Roberts, K.~Lee, S.~Narang, M.~Matena, Y.~Zhou,
  W.~Li, and P.~J. Liu\leavevmode\nopagebreak\newline 2019.
\newblock Exploring the limits of transfer learning with a unified text-to-text
  transformer.
\newblock {\em arXiv preprint arXiv:1910.10683}.

\bibitem[\protect\astroncite{Santoro et~al.}{2016}]{MANN}
Santoro, A., S.~Bartunov, M.~Botvinick, D.~Wierstra, and T.~P.
  Lillicrap\leavevmode\nopagebreak\newline 2016.
\newblock One-shot learning with memory-augmented neural networks.
\newblock {\em CoRR}, abs/1605.06065.

\bibitem[\protect\astroncite{Socher et~al.}{2013}]{sst2}
Socher, R., A.~Perelygin, J.~Wu, J.~Chuang, C.~D. Manning, A.~Ng, and
  C.~Potts\leavevmode\nopagebreak\newline 2013.
\newblock Recursive deep models for semantic compositionality over a sentiment
  treebank.
\newblock In {\em Proceedings of the 2013 conference on empirical methods in
  natural language processing}, Pp.~ 1631--1642.

\bibitem[\protect\astroncite{Srivastava et~al.}{2014}]{dropout}
Srivastava, N., G.~Hinton, A.~Krizhevsky, I.~Sutskever, and
  R.~Salakhutdinov\leavevmode\nopagebreak\newline 2014.
\newblock Dropout: A simple way to prevent neural networks from overfitting.
\newblock {\em Journal of Machine Learning Research}, 15:1929--1958.

\bibitem[\protect\astroncite{Wang}{2018}]{DRNN}
Wang, B.\leavevmode\nopagebreak\newline 2018.
\newblock Disconnected recurrent neural networks for text categorization.
\newblock In {\em Proceedings of the 56th Annual Meeting of the Association for
  Computational Linguistics (Volume 1: Long Papers)}, Pp.~ 2311--2320.

\bibitem[\protect\astroncite{Yang et~al.}{2019}]{XLNet}
Yang, Z., Z.~Dai, Y.~Yang, J.~G. Carbonell, R.~Salakhutdinov, and Q.~V.
  Le\leavevmode\nopagebreak\newline 2019.
\newblock Xlnet: Generalized autoregressive pretraining for language
  understanding.
\newblock {\em CoRR}, abs/1906.08237.

\end{thebibliography}

\newpage
\appendix
\section{Appendix}
\subsection{Input Representation and Training Details}
\label{mcq}
\subsubsection{Input Tokens}
\begin{figure*}[h!]
	\begin{center}
		\includegraphics[scale=.45]{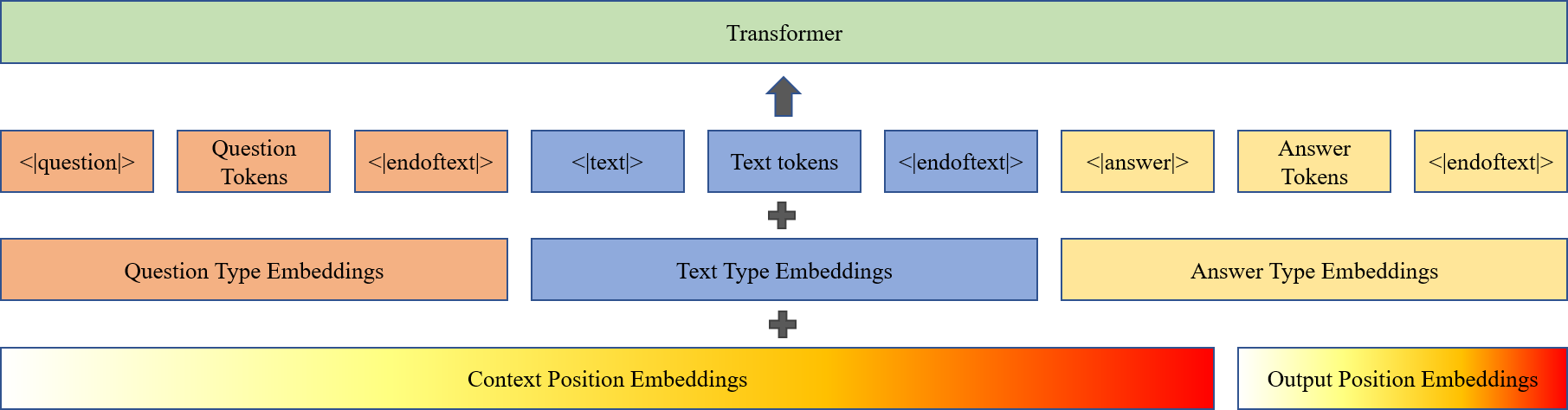}
		\label{fig:input_rep}
	\end{center}
\end{figure*}

To form the input representation, the question, text, and answer tokens are concatenated together. Each set of tokens has a \texttt{<|endoftext|>} token appended to the end, and has a special prompt token prepended to the set. The special tokens for the three fields are respectively \texttt{<|question|>}, \texttt{<|text|>}, and \texttt{<|answer|>}. In addition to prompt tokens, each segment of the input also has unique type token embeddings added. There are three different type tokens total, one for each segment of the input. Lastly, to encode positional information in our input representation we utilize two sets of position embeddings. One range of position ids up to and including the \texttt{<|answer|>} prompt token, and another set of ids starting from 0 at the beginning of the answer tokens. These ranges are depicted by the colored gradient in the figure above. This helps the transformer distinguish between the context and the generated output.
\subsubsection{Multiple Choice Format}

We maintain a list of approximately 25 multiple choice question formats as shown below. At training and evaluation time we randomly sample a question format and fill the brackets with the desired classes. We format the classes as a comma separated list with double quotation marks to help segment the answers from the rest of the question text. We ensure that spaces are put between the answers and the quotation marks to avoid any unwanted byte pair merges:  \texttt{`` class1 '' , `` class2 '' , or `` class3 ''}. Examples of this formatting can be seen in Table \ref{tab:example_data}.

\begin{itemize}
	\item To which category does the following document belong? : \{\}
	\item To which category does the following text belong? : \{\}
 	\item To which category does the text belong? : \{\}
	\item To which category does the article belong? : \{\}
	\item How would you describe the following document? : as \{\} 
	\item How would you describe the text? : as \{\}
	\item How would you describe the following text? : as \{\}
	\item Which best describes the text? : \{\}
	\item Which best describes the document? : \{\}
	\item Which best describes the following document? : \{\}
	\item Which best describes the following text? : \{\}
	\item The following document is \_ ? : \{\}
	\item The following text is \_ ? : \{\}
	\item The text is \_ ? : \{\}
	\item The document is \_ ? : \{\}
	\item How is the text best described? : \{\}
	\item How is the document best described? : \{\}
	\item How is the following text best described? : \{\}
	\item How is the following document best described? : \{\}
	\item Which of these choices best describes the text? : \{\}
	\item Which of these options best describes the text? : \{\}
	\item Which of these choices best describes the document? : \{\}
	\item Which of these options best describes the document? : \{\}
	\item Which of these categories best describes the following document? : \{\}
	\item Which of these choices best describes the following document? : \{\}
	\item Which of these options best describes the following text? : \{\}
\end{itemize}

\subsection{Training Hyperparameters}
To train our model we follow a procedure largely based on the training procedures described in \cite{Radford2019GPT2} with a few differences. All training is performed with a maximum sequence length of 512 tokens. In the full dataset training setting we utilize a learning rate of $4\times10^{-5}$ and a batch size of 128. When training with a quarter of the dataset we then used a learning rate of $3\times10^{-5}$ and a batch size of 32. Our learning rate has a warmup period over 1\% of the total training iterations before decaying according to a single cycle cosine decay schedule over 10 epochs. We utilize an Adam optimizer \citep{Adam} with decoupled weight decay \citep{loshchilov2018decoupled} $\lambda = 0.01$. All our models are trained efficiently on V100 GPUs by utilizing mixed precision training with dynamic loss scaling \citep{MPTraining}. Additionally, we use global gradient norm clipping of 1.0 to improve the stability of training large models. Lastly, we utilize attention and hidden state dropout \citep{dropout} values of 0.1.

\subsection{Training Data Statistics}
\label{data_statistics}
We provide class frequency statistics shown below to highlight the diversity of the dataset used for pretraining. 
\begin{figure*}[h!]
	\begin{center}
		\includegraphics[scale=.7]{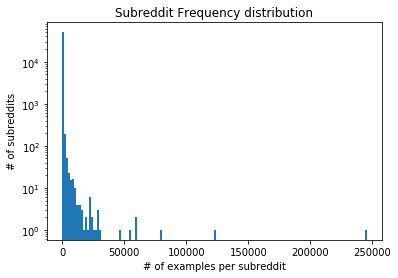}
		\caption{Subreddit Class Distribution. The number of times a subreddit occurs (frequency) is presented on the x-axis. The y-axis corresponds to the number of subreddits that appear at a certain frequency.}
		\label{fig:data_distro}
	\end{center}
\end{figure*}

The data is distributed according to a power law distribution clustered around <1000 samples per subreddit, with a long tail reaching up to 245000 samples for a given subreddit. Zooming into the distribution (shown below) we find that there are approximately 9400 subreddits with 20 or more samples out of 50700 subreddit. Out of the 9400 subreddit two thirds have fewer than 100 samples. This level of diversity is ideal for a meta learning or domain adaptation dataset.

\begin{figure*}[h!]
	\begin{center}
		\includegraphics[scale=.7]{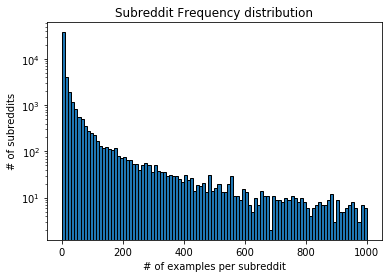}
		\caption{Enlarged Subreddit Class Distribution.}
		\label{fig:data_distro_zoom}
	\end{center}
\end{figure*}

Lastly we show the most common subreddits along with their subreddit frequency in Table \ref{tab:top15}. We find that half of the top fifteen subreddits are politically related. This skew may lead to possible biases in the training process. A plausible explanation for this bias can be found in the way the dataset is collected. Since we heuristically filter for reputable outbound links it is likely that we choose subreddits where people post outside news.

\begin{table*}[h!]
	\footnotesize
	\begin{center}
		\label{tab:top15}
		\begin{tabular}{|c|c|}
			\hline 
			\thead{Subreddit} & \thead{Frequency} \\ \hline
			r/politics & 245308 \\\hline
			r/worldnews & 122884 \\\hline
			r/The\_Donald & 80042 \\\hline
			r/todayilearned & 59892 \\\hline
			r/news & 59166 \\\hline
			r/technology & 54860 \\\hline
			r/science & 46452 \\\hline
			r/Conservative & 30823 \\\hline
			r/POLITIC & 28310 \\\hline
			r/conspiracy & 28293 \\\hline
			r/india & 27892 \\\hline
			r/environment & 26816 \\\hline
			r/atheism & 25999 \\\hline
			r/programming & 24020 \\\hline
			r/Libertarian & 23711 \\\hline
		\end{tabular}
		\caption{Subreddit Frequency.}
	\end{center}
\end{table*}

\newpage
\subsection{Downstream Task Setup}
\subsubsection{Class Descriptors}
\label{class_description}
Listed below are the class descriptions used for each classification task.

\begin{table*}[h!]
	\footnotesize
	\begin{center}
		\label{tab:class_description}
		\begin{tabular}{|c|p{11cm}|}
			\hline 
			\thead{Dataset} & \thead{Classes} \\ \hline
			SST-2 & Positive Sentiment, Negative Sentiment \\\hline
			AGNews & Science \& Technology, Business, Sports , World News \\\hline
			DBPedia & Company, Mean Of Transportation, Film, Office Holder, Written Work, Animal, Natural Place, Artist, Plant, Athlete, Album, Building, Village, Educational Institution \\\hline
			Yahoo Answers & Family \& Relationships, Business \& Finance, Health, Society \& Culture, Education \& Reference, Entertainment \& Music, Science \& Mathematics, Computers \& Internet, Sports, Politics \& Government
			\\\hline
			Yelp-2 & Positive polarity, Negative polarity \\\hline
			Amazon-2 & Positive polarity, Negative polarity \\
			\hline
		\end{tabular}
	\end{center}
\end{table*}

\subsubsection{Descriptor Selection}
\label{qsec}
The ability of our model to adapt to new tasks and its behavior for a given input is controlled by the input descriptor questions it receives. In this section we investigate the impact that question formulation has on downstream task performance. Specifically, we modify the provided class descriptions for several tasks and observe the effects this has on the 355 million parameter model's downstream task performance: 
\begin{itemize}
	\item For binary classification tasks like SST-2, Amazon-2, Yelp-2 we move away from \texttt{Positive Sentiment} and \texttt{Negative Sentiment}, or \texttt{Positive polarity} and \texttt{Negative polarity}. Instead we simply use \texttt{positive} and \texttt{negative} as in \cite{mqan}.
	\item For DBPedia we revert to the original class descriptions provided by the dataset and remove all whitespace (eg. \texttt{Mean Of Transportation} becomes \texttt{MeanOfTransportation}). 
	\item For AGNews we also revert to the original class descriptions and change \texttt{World News} to \texttt{World} and \texttt{Science \& Technology} to \texttt{Sci/Tech}.
\end{itemize}

Table \ref{tab:tokenization_results} shows that the choice of class description has a significant impact on performance. In the worst case poor class descriptions can lead to an absolute 27\% drop in accuracy and 44\% increase in out of vocabulary answers. In the cases of binary classification tasks and AGNews we hypothesize performance is negatively impacted by incomplete task descriptions: \texttt{positive} and \texttt{World} do not explicitly convey positive sentiment or World News. Empirical observations in Figure \ref{fig:confusions} show that the model either selects plausibly overlapping categories in the case of AGNews, or responds with a completely out of vocabulary answer as in the case of sentiment analysis. For DBPedia and AGNews, concatenating words together drastically changes the resulting bytepair tokenization despite the descriptions still being human readable. This changes the semantic understanding that the model receives and as a result the model completely avoids selecting it. In some cases the model may not have ever trained the subword embeddings corresponding to those tokens. This section highlights that our language modeling technique, while general, is subject to errors arising from problem formulation and requires careful control to craft questions that elicit desired effects. Remedying these issues will be a goal of future work.
\begin{table}[h!]
\footnotesize
\begin{center}

\label{tab:tokenization_results}
\begin{tabular}{|c|c|c|c|c|c|}
\hline
Descriptor Set & SST-2 & AGNews & DBPedia & Amazon-2 & Yelp-2 \\ \hline
Good Descriptors & 63.22 / 0 & 69.04 / .478 & 53.85 / .056 &  81.22 / .056 & 74.35 / 0 \\\hline
Bad Descriptors & 35.91 / 44.3 & 62.61 / 0 & 44.99 / .050 & 64.3 / 22.1 & 68.02 / 23.4 \\\hline

\end{tabular}
\vspace{2mm}
\caption{Validation Set Accuracy/Out of Vocabulary Answer Percentages. We compare performance on the validation set with two different sets of descriptors: one deemed good and one deemed bad. We showcase the importance of selecting appropriate descriptors for a task.}

\end{center}
\end{table}

\begin{figure*}[ht!]
	\begin{center}
		\includegraphics[scale=.8]{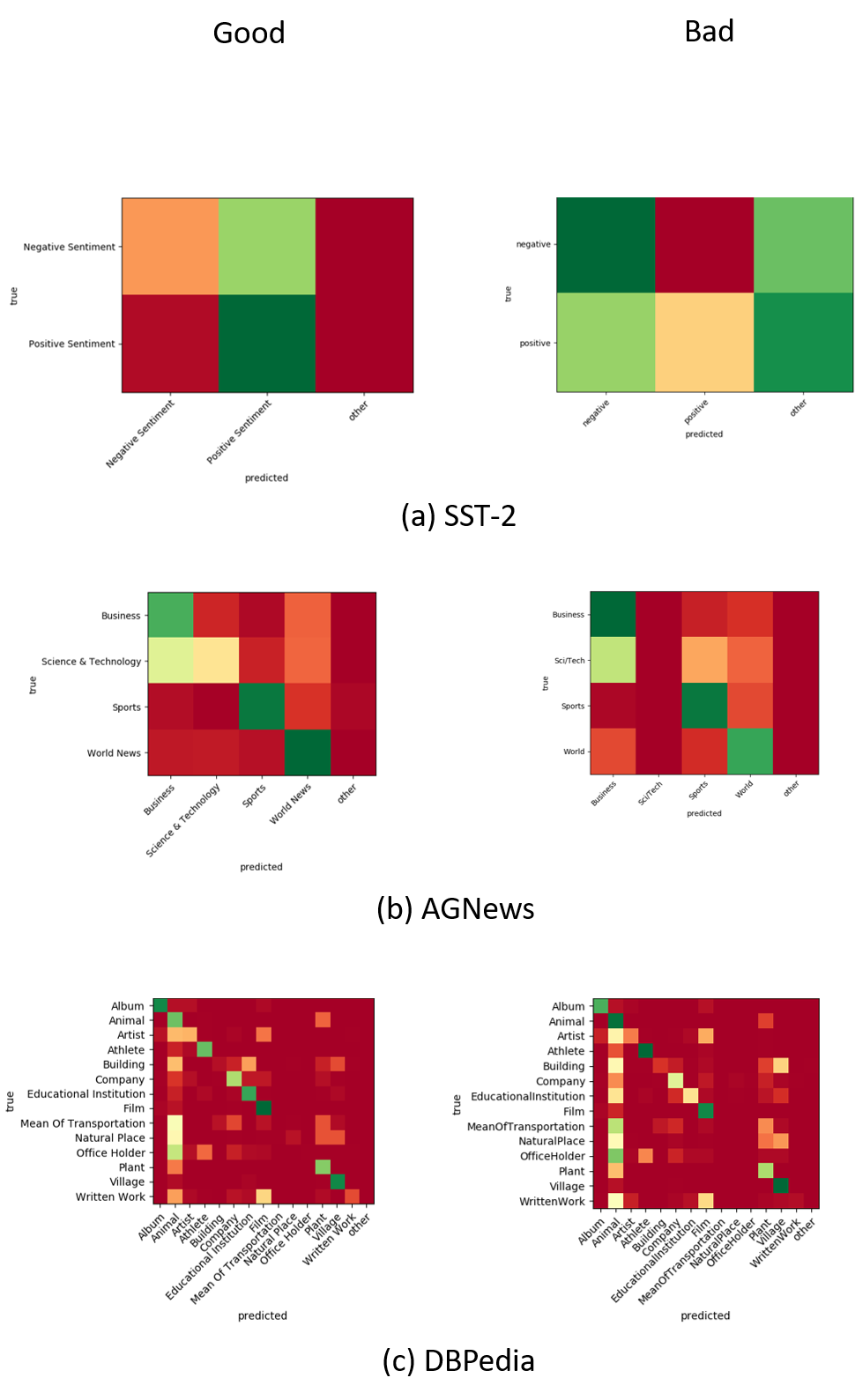}
		\caption{Confusion matrices for several classification tasks. The left column corresponds to the first row in Table \ref{tab:tokenization_results}, and the right column corresponds to the second row. The color represents the prediction frequency with green being the highest, red the lowest, and yellow in the middle.}
		\label{fig:confusions}
	\end{center}
\end{figure*}

\end{document}